%% file: arxiv.tex
\documentclass{article}
\usepackage{times}
\usepackage{hyperref}
\usepackage{url}
\usepackage{fullpage}

\usepackage{amsmath}
\usepackage{subfig}
\usepackage{enumitem}

\usepackage{tikz}
\usepackage{tikz-qtree}
\usetikzlibrary{patterns}
\usetikzlibrary{shapes}
\usetikzlibrary{arrows}

\usepackage{grfext}
\PrependGraphicsExtensions*{.pdf}

\newcommand{\cw}{\tilde{w}}
\newcommand{\given}{\,|\,}

\title{Graph-Sparse LDA: A Topic Model \\ with Structured Sparsity}

\author{
Finale Doshi-Velez \\ 
Harvard University \\
finale@seas.harvard.edu
\and
Byron C. Wallace \\ 
University of Texas at Austin \\ 
byron.wallace@utexas.edu
\and 
Ryan P. Adams \\ 
Harvard University \\
rpa@seas.harvard.edu
}

\begin{document} 

\maketitle

\begin{abstract}
Originally designed to model text, topic modeling has become a
powerful tool for uncovering latent structure in domains including
medicine, finance, and vision.  The goals for the model vary depending
on the application: in some cases, the discovered topics may be used
for prediction or some other downstream task.  In other cases, the
content of the topic itself may be of intrinsic scientific interest.

Unfortunately, even using modern sparse techniques, the discovered
topics are often difficult to interpret due to the high dimensionality
of the underlying space.  To improve topic interpretability, we
introduce Graph-Sparse LDA, a hierarchical topic model that leverages
knowledge of relationships between words (e.g., as encoded by an
ontology). In our model, topics are summarized by a few latent
\emph{concept-words} from the underlying graph that explain the
observed words. Graph-Sparse LDA recovers sparse, interpretable
summaries on two real-world biomedical datasets while matching
state-of-the-art prediction performance.

\end{abstract} 

\section{Introduction}
\label{section:intro}

Probabilistic topic models \cite{blei-03,steyvers-07,blei-12} were
originally developed to discover latent structure in unorganized text
corpora, but these models have been generalized to provide a powerful
and flexible framework for uncovering structure in a variety of
domains including medicine, finance, and vision.  In the popular
Latent Dirichlet Allocation (LDA) \cite{blei-03} model, \emph{topics}
are distributions over the words in the vocabulary, and documents can
then be summarized by the mixture of topics they contain.  Here, a
``word'' is anything that can be counted and a ``document'' is an
observation.  LDA has been applied to diverse applications such as
finding scientific topics in articles \cite{griffiths-04a},
classifying images \cite{fei-05}, and recognizing human actions
\cite{wang-09a}.  The modeling objective varies depending on the
application.  In some cases, topic models are used to provide compact
summaries of documents which can then be used for downstream tasks
such as prediction, classification, or recognition.  In other
situations, the content of the topics themselves may be of independent
interest.  For example, a clinician may want to understand \emph{why}
a certain topic within their patient's data is correlated with
mortality (e.g., \cite{ghassemi}). A geneticist, meanwhile, may wish
to use topics discovered from publicly available datasets to formulate
the next hypothesis to be tested in an expensive laboratory study.

These kinds of applications present unique challenges and opportunities
for topic modeling.  In the standard LDA formulation, topics are
distributions over all of the words in a (usually very large)
vocabulary.  This vocabulary is typically assumed to be unstructured,
i.e., words are not assumed to have any \emph{a priori} relationship.
Sparse topic models \cite{archambeau-11,williamson-10,eisenstein-11}
offer a partial solution to this problem by enforcing the constraint
that many of the word probabilities for a given topic should be zero.  
Unfortunately, when the vocabularies are large, there may still 
be hundreds of words with non-zero probabilities. Enforcing sparsity 
alone is therefore not sufficient to induce interpretable topics.

In this work we propose a new strategy for achieving interpretability:
exploiting structured vocabularies, which exist in many specialized
domains. These ``controlled structured vocabularies'' encode known
relationships between the tokens comprising the vocabulary.  For
example, diseases are organized into billing hierarchies, and clinical
concepts are related by directed acyclic graphs
(DAGs)~\cite{bodenreider-04}. There are other examples as well.
Keywords for biomedical publications are organized in a hierarchy
known as MeSH \cite{lipscomb-00}; searching with MeSH terms is
standard practice for biomedical literature retrieval tasks.  Genes
are organized into pathways and interaction networks.  Such structures
often summarize large bodies of scientific research and human thought;
a great deal of effort has gone into their construction.  While these
structured vocabularies are necessarily imperfect, they have the
important property that they --- by definition --- represent how
\emph{domain experts codify knowledge}, and thus provide a window into
how one might create models that such experts can meaningfully use and
interpret.  Because they were designed to be understood by humans,
these structured relationships provide a form of information unique
from any learned ontology.

\begin{figure}[t]
\centering
\begin{minipage}[t]{0.48\textwidth}
\includegraphics[width=\textwidth]{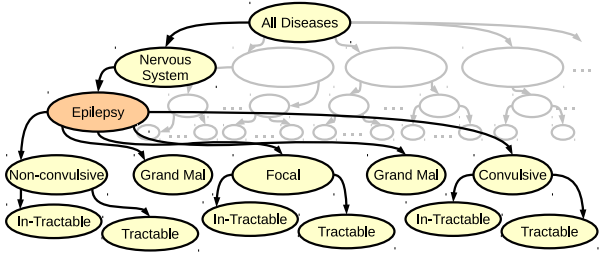}
\caption{Simplified section of the ICD9-CM diagnostic hierarchy.
  Here, ``Epilepsy'' might be a good \emph{concept-word} to
  summarize the very specific forms of epilepsy that are its
  descendants. Knowing that a patient has epilepsy may also explain
  instances of ``Central Nervous System Disorder'' or even
  ``Disease.''}
\label{fig:icd-example}
\end{minipage}\hfill
\begin{minipage}[t]{0.50\textwidth}
\resizebox{\textwidth}{!}{%
\begin{tikzpicture}[
    every tree node/.style={draw,align=center,thick,font=\scriptsize,circle, minimum height=15pt, minimum width=15pt},
    edge from parent/.style={draw,edge from parent path={(\tikzparentnode.south)--+(0,-2pt)-| (\tikzchildnode)}},
    level distance=20pt,
    sibling distance=5pt,
    level 1/.style={sibling distance=1pt},
    level 2/.style={sibling distance=1pt},
    level 3/.style={sibling distance=1pt},
    level 4/.style={sibling distance=1pt},
    edge from parent path={(\tikzparentnode) -- (\tikzchildnode)},
    edge from parent/.append style={=>,thick}
    ]
\Tree
[.\node[fill=brown!50] {};
    [.\node[fill=red!30] {};
        [ .\node {}; 
            [.\node {};
                [.\node {};]
                [.\node {};]
            ]
            [.\node {};
                [.\node {};]
                [.\node {};]
            ]
        ]
        [ .\node[fill=red!30] {}; 
            [.\node[fill=red!100] {$1$};
                [.\node[ellipse,fill=red!30] {};]
                [.\node[ellipse,fill=red!30] {};]
            ]
            [.\node {};
                [.\node {};]
                [.\node {};]
            ]
        ]
    ]
    [.\node[fill=green!50] {};
        [ .\node[fill=yellow!100] {$2$}; 
            [.\node[fill=yellow!30] {};
                [.\node[fill=yellow!30] {};]
                [.\node[fill=yellow!30] {};]
            ]
            [.\node[fill=yellow!30] {};
                [.\node[fill=yellow!30] {};]
                [.\node[fill=yellow!30] {};]
            ]
        ]
        [ .\node[fill=cyan!100] {$3$}; 
            [.\node[fill=cyan!30] {};
                [.\node[fill=cyan!30] {};]
                [.\node[fill=cyan!30] {};]
            ]
            [.\node[fill=cyan!30] {};
                [.\node[fill=cyan!30] {};]
                [.\node[fill=cyan!30] {};]
            ]
        ]
    ]
]
\end{tikzpicture}
}
\caption{Example tree structure, where every node (including interior nodes)
represents a vocabulary word.  A concept-word can explain instances of its
descendants and ancestors, e.g., if node 1 is a concept word, the matrix $P$
would only have non-zero values for the descendants and ancestors, marked in red and brown.
}
\label{fig:toy_cartoon}
\end{minipage}
\vspace{-1.8em}
\end{figure}

Unfortunately, existing topic modeling machinery is not equipped 
to capitalize on controlled structured vocabularies. We therefore 
propose a new model, Graph-Sparse LDA, that exploits 
DAG-structured vocabularies to induce interpretable topics
that still summarize the data well.  This approach is appropriate when
documents come annotated with structured vocabulary terms, e.g.,
biomedical articles with MeSH headers, genes with known interactions,
and species with known taxonomies.  Graph-Sparse LDA introduces an
additional layer of hierarchy into the standard LDA model: instead of
topics being distributions over observed words, topics are
distributions over \emph{concept-words}, which 
then generate observed words using a noise process that is
informed by the structure of the vocabulary (see example in
figure~\ref{fig:icd-example}).  Using the structure of the
vocabulary to guide the induced sparsity, we recover topics
that are more interpretable to domain experts.


We demonstrate Graph-Sparse LDA on two real-world applications.  The
first is a collection of diagnoses for patients with autism spectrum
disorder. For this we use a diagnosis hierarchy \cite{bodenreider-04}
to recover clinically relevant subtypes described by a small set of
concepts.  The second is a corpus of biomedical abstracts annotated
with hierarchically-structured Medical Subject Headings
(MeSH)~\cite{lipscomb-00}.  Here, Graph-Sparse LDA identifies
meaningful, concise groupings (topics) of MeSH terms for use in
biomedical literature retrieval tasks.  In both cases, the topic
models found by Graph-Sparse LDA have the same or better predictive
performance as a state-of-the-art sparse topic model (Latent IBP
compound Dirichlet Allocation \cite{archambeau-11}) while providing
much sparser topic descriptions.  To efficiently sample from this
model, we introduce a novel inference procedure that prefers moves
along manifolds of constant likelihood to identify sparse solutions.

\section{Graph-Sparse LDA}
\label{section:methods}

In this paper, our data are documents that are modeled using the ``bag of words'' representation that is common for topic models.
Let the data $X$ consist of the counts of each of the $V$ words in the
vocabulary for each of the $N$ documents.  The standard LDA model
\cite{blei-03} posits the following generative process for the words
$w_{in}$ comprising each document (data instance) in $X$:
\begin{align}
  B_n &\sim \textrm{Dirichlet}( \alpha_B 1_K) &
  A_k &\sim \textrm{Dirichlet}( \alpha_A 1_V) \\
  z_{in} \given B_n &\sim \textrm{Discrete}( B_n ) &
  w_{in} \given z_{in} , \{ A_k \} &\sim \textrm{Discrete}( A_{z_{in}} )  
\end{align}
where $K$ is the number of topics.  The rows of the~${N\! \times\! K}$
matrix $B$ are the document-specific distributions over topics, and
the~${K\! \times\! V}$ matrix $A$ represents each topic's distribution
over words.  The notation~$A_k$ refers to the~$k^{th}$ row of~$A$
and~$B_n$ is the~$n^{th}$ row of~$B$.  The $z_{in}$ encode the topic
to which the $i^{th}$ word in document $n$ was assigned, and
${w_{in}\! \in 1,\ldots,V}$ is the~$i^{th}$ word in document~$n$.
Since the words are assigned independently and identically, an~${N
  \times V}$ matrix of how often each word occurs in each document is
a sufficient statistic for the words~$w_{in}$.

Our Bayesian nonparametric model, Graph-Sparse LDA, builds upon a recent nonparametric extension of LDA, Latent IBP compound Dirichlet Allocation (LIDA) \cite{archambeau-11}.  In addition to allowing an unbounded number of topics, LIDA introduces sparsity over both the document-topic matrix~$B$ and the topic-word matrix~$A$ using a three-parameter Indian Buffet Process.  The prior expresses a preference for describing each document with few topics and each topic with few words.
We extend LIDA by assuming that words in our document
belong to a structured vocabulary with known relationships that form a
tree or DAG, and that nearby groups of
terms---as defined with respect to the graph structure---are
associated with specific phenomena.  For example, in a biomedical
ontology, nodes on one sub-tree may correspond to a particular virus
(e.g., HIV) and a different sub-tree may describe a specific drug or
treatment (e.g., anti-retrovirals) used to treat HIV.  Papers
investigating anti-retrovirals for treatment of HIV would then tend to
have terms drawn from both sub-trees. Intuitively, we would like to
uncover these sub-trees as the concepts underpinning a topic.

Using concept-words to summarize the words in a topic is natural in
many scenarios because structured vocabularies are often both very
specific and inconsistently annotated.  For example, a trial may be
annotated with the term \emph{antiviral agents} or its child
\emph{anti-retroviral agents}.  Thus, from a generative modeling
perspective, nearby words in the vocabulary can be thought of as
having been produced from the same core concept.  Our model posits
that a topic is made up of a sparse set of concept-words that can
explain words that are its ancestors or descendants (see
Figure~\ref{fig:icd-example}).  Formally, we replace the previous LDA generative process with the following process that introduces~$\cw_{in}$ as the concept word behind observed word~$w_{in}$:
\begin{align}
  \pi_k &\sim \textrm{IBP-Stick}( \gamma_B ) &
  \rho_v &\sim \textrm{Beta}( \gamma_A/V , 1 ) \\
  \bar{B}_n \given \pi_k &\sim \textrm{Bernoulli}( \pi_k )  &
  \bar{A}_k \given \rho_v &\sim \textrm{Bernoulli}( \rho_v ) \\
  B_n \given \bar{B}_n &\sim \textrm{Dirichlet}( \bar{B}_n \odot \alpha_B 1_K ) &
  A_k \given \bar{A}_k &\sim \textrm{Dirichlet}( \bar{A}_k \odot \alpha_A 1_V ) \\
  z_{in} \given B_n &\sim \textrm{Discrete}( B_n ) &
  \cw_{in} \given z_{in} , \{ A_k \} &\sim \textrm{Discrete}( A_{z_{in}} ) \\
  P_v &\sim \textrm{Dirichlet}( \mathcal{O}_v \odot \alpha_P 1_V ) &
  w_{in}\given \cw_{in},P &\sim \textrm{Discrete}( P_{\cw_{in}} )
\end{align}
where $\odot$ is the element-wise Hadamard product and IBP is the Indian Buffet Process \cite{griffiths-11}.  As in the standard LDA model, the document-topic matrix $B$ represents the distribution of topics in each document.  However,~$B_n$ is now masked according to a document-specific vector~$\bar{B}_n$, which is the~$n^{th}$ row of a matrix~$\bar{B}$ that is itself drawn from an IBP with concentration parameter~$\gamma_B$.  Thus $\bar{B}_{nk}$ is 1 if topic~$k$ has nonzero probability in document~$n$ and 0 otherwise.  Similarly, the topic-concept matrix~$A$ and the binary topic-concept mask matrix $\bar{A}$ represent the topic matrix and its sparsity pattern, except that now~$A$ and~$\bar{A}$ represent the relationship between topics and concept-words.  The priors over the document-topic and topic-concept matrices~$B$ and~$A$ (and their respective masks~$\bar{B}$ and~$\bar{A}$) follow those in LIDA~\cite{archambeau-11}.

The concept-word matrix $P$ describes distributions over words for
each concept.  The form of the ontology $\mathcal{O}$ determines the
sparsity pattern of $P$: we use the notation $\mathcal{O}_{w}$ to
refer to a binary vector of length $V$ that is 1 if the concept-word
$\cw$ is a descendant or ancestor of observed word $w$ and 0
otherwise.  We illustrate these sparsity constraints in
Figure~\ref{fig:toy_cartoon}, where the dark-shaded concept nodes 1,
2, and 3 can each only explain themselves, and words that are are
ancestors or descendants.  The brown and green nodes are ancestor
observed words that are shared by more than one concept word.

Intuitively, the concept-word matrix $P$ can be viewed as allowing for
variation in the process of assigning terms to documents (citations,
diagnoses, etc.) on behalf of domain experts.  For example, if a
document is about \emph{anti-retroviral agents}, an annotator may
describe the document with a key-word nearby in the vocabulary, such
as \emph{antiviral agents}, rather than the more specific term.
Similarly, a primary care physician using the hierarchy in
Figure~\ref{fig:icd-example} may note that a patient has
\emph{epilepsy} since he is not an expert in neurological disorders,
while a specialist might bill for the more specific term
\emph{Convulsive Epilepsy, Intractable}.  More generally, the
concept-word matrix~$P$ can be thought of as describing a neighborhood
of words that could be covered by the same concept.  Introducing this
additional layer of hierarchy allows us to find very sparse
topic-concept matrices $A$ that still explain a large number of
observed words.  (Note that setting~${P = I_V}$ recovers LIDA from
Graph-Sparse LDA; Graph-Sparse LDA is therefore a generalization of LIDA
that allows for much more structure.)

Finally, let the data $X$ be an ${N \times V}$ matrix of counts, where
$X_{nw}$ is the number of times word $w$ appears in
 document $n$.  The log-likelihood of the data given $B$, $A$, and $P$
 is given by
 \begin{align}
 \log p(X | A, B, P) &= \sum_{n,\cw} X_{n\cw} \log( B_n A P_{\cw} )
 \label{eqn:ll}
 \end{align}

\section{Inference}
We describe a blocked-Gibbs procedure for
sampling~$B$,~$\bar{B}$,~$A$,~$\bar{A}$, and~$P$ as well as an
additional Metropolis-Hastings (MH) procedure that helps the sampler
to move toward sparser topic-concept word matrices~$A$.  Specifically,
our MH proposal distribution is designed to prefer proposals of
new~$A'$ and~$P'$ such that the overall likelihood does not
significantly change.  To our knowledge, MCMC that uses moves that
result in near-constant likelihood to encourage large changes in the
prior is a novel approach.  We first describe how to resample
instantiated parameters of the Graph-Sparse LDA model and then
describe how we sample new topics.

\subsection{Blocked Gibbs Sampling} 
Our blocked Gibbs sampling procedure relies on first sampling two
intermediate assignment tensors.  The first,~$C_{NKV}$ counts how
often the word~$v$ is assigned to topic $k$ in document $n$.  The
second, $C_{KVV}$ counts how often each observed word is assigned to
each concept word in topic~$k$.  These two tensors are sampled as
follows:

\paragraph{Count Tensors $C_{NKV}$ and $C_{KVV}$:}
The probability that an observed word~$w$ belongs to topic $k$ is
given by ${p_{n k w} = B_{nk} \sum_{\cw=1}^V A_{k\cw} P_{\cw w}}$
where the sum over $\cw$ marginalizes out potential
concept-words. Thus we can use a multinomial distribution to allocate
the counts~$X_{nw}$ across the~$K$ topics via~${[C_{NKV}]_{n:w} \sim
  \textrm{Mult}(p_{n:w}, X_{nw})}$, where we use ``:'' to indicate a
tensor slice.  For updating $C_{KVV}$, the probability that~$\cw$ was
the generating concept word, given the observed word~$w$ and the topic
$k$, is given by ${p_{k \cw w} \propto A_{k\cw} P_{\cw w}}$.  Thus we
can sample the count tensor~$C_{KVV}$ using the multinomial
\begin{equation*}
{[C_{KVV}]_{k:w} \sim \textrm{Mult}( p_{k : w} , {\textstyle \sum_n
    [C_{KNV}]_{nkw}})}
\end{equation*}  
Note that given the topic assignment for an observed word, we do not
need to know from which document it came to determine the distribution
over concept-word assignments.  Thus, we never need to consider a
four-way \{document, topic, concept-word, word\} count tensor during
inference.

\paragraph{Document-Topic Assignments $B$ and $\bar{B}$:}  
Given the count tensor $C_{NKV}$, we can sample the sparsity mask
$\bar{B}$ by marginalizing out~$B$ and the~$\pi_k$ using the formula
derived in \cite{archambeau-11}.  First, we note that if
${\sum_{w=1}^V[C_{NKV}]_{nkw} > 0}$, then $\bar{B}_{nk}$ must be 1
because there exists at least one word assigned to topic $k$ in
document $n$.  Let~$\bar{B}^{(nk=0)}$ denote the matrix~$\bar{B}$, but
with entry ${\bar{B}_{nk}=0}$.  If ${\sum_{w=1}^V [C_{NKV}]_{nkw} =
  0}$, then the probability $\phi_{nk}$ that ${\bar{B}_{nk} = 1}$ is
given by
\begin{align*}
  \psi_{nk}&=\left[1+
    \frac{\mathcal{B}( \alpha_B \bar{B}^{(nk=0)}_n, \alpha_B)    
    ( N - \bar{B}^{(nk=0)}_k) }
    { \mathcal{B}( \alpha_B \bar{B}^{(nk=0)}_n + \sum_{w=1}^V X_{nw}, \alpha_B ) 
     \bar{B}^{(nk=0)}_k }\right]^{-1}
\label{eqn:bbar}
\end{align*}
where $\mathcal{B}(\cdot,\cdot)$ is the Beta function,  ${\bar{B}^{(nk=0)}_n \equiv \sum_{k'}\bar{B}^{(nk=0)}_{nk'}}$, and ${\bar{B}^{(nk=0)}_k \equiv \sum_{n'}\bar{B}^{(nk=0)}_{n'k}}$. Once we have resampled~$\bar{B}$, we resample $B$ using 
\begin{equation*}
{B_n \sim \textrm{Dirichlet}( \bar{B}_n \odot ( \alpha_B + {\textstyle \sum_{w=1}^V [C_{NKV}]_{n:w}}) )}.
\end{equation*}

\paragraph{Topic-Concept Word Assignments $A$ and $\bar{A}$:}  
As with~$\bar{B}$, we marginalize out $A$ and $\rho_v$ when sampling $\bar{A}$.  If ${\sum_{w=1}^V[C_{KVV}]_{k \cw w} > 0}$, then at least one observed
word was assigned to topic $k$ and concept-word $\cw$, and therefore ${\bar{A}_{k\cw} = 1}$.  Let~$\bar{A}^{(k\cw=0)}$ be the same as matrix~$\bar{A}$, but with entry~${\bar{A}_{k\cw}=0}$. If ${\sum_{w=1}^V[C_{KVV}]_{k\cw w} = 0}$, then the probability $\phi_{k\cw}$ that~${\bar{A}_{k\cw} = 1}$ is given by
\begin{align*}
\phi_{k\cw} &= 
\left[1 +
    \frac{\mathcal{B}( \alpha_A \bar{A}^{(k\cw=0)}_k, \alpha_A ) 
    ( K - \bar{A}^{(k\cw=0)}_{\cw} )) }{
    (\mathcal{B}( \alpha_A \bar{A}^{(k\cw=0)}_k +\sum_{\cw}\sum_w[C_{KVV}]_{k\cw w}, \alpha_A )
    \bar{A}^{(k\cw=0)}_{\cw}}\right]^{-1}
\end{align*}
where $\mathcal{B}$ is the Beta function,~$K$ is the number of instantiated topics, 
${\bar{A}^{(k\cw=0)}_k \equiv \sum_{\cw'}\bar{A}^{(k\cw=0)}_{k\cw'}}$, and
${\bar{A}^{(k\cw=0)}_{\cw} \equiv \sum_{k'}\bar{A}^{(k\cw=0)}_{k'\cw}}$.  Once we have resampled $\bar{A}$, we resample $A$ via 
\begin{equation*}
{A_k \sim \textrm{Dirichlet}( \bar{A}_k \odot ( \alpha_A +
  {\textstyle\sum_{w}[C_{KVV}]_{k:w}}))}
\end{equation*}

\paragraph{Concept Word-Word Distributions $P$:}  
Finally, the concept word to observed word distributions can be resampled via
\begin{equation*}
{P_{\cw} \sim \textrm{Dirichlet}( ( \mathcal{O}_{\cw} > 0 ) \odot ( \alpha_P + {\textstyle \sum_k [C_{KVV}]_{k\cw:}}))}.
\end{equation*}

\subsection{MH Moves for Improved Sparsity}
Recall that one of our modeling objectives is to identify a small,
interpretable set of concept-words in each topic. To this end, we have
placed a sparsity-inducing prior on $A$.  While the Gibbs sampling
procedure above is computationally straightforward, it often does not
give us the desired sparsity in~$A$ fast enough.  Mixing is slow
because the only time we set~${\bar{A}_{k\cw} = 0}$ is when no counts
of~$\cw$ are assigned to topic~$k$ \emph{across any of the documents}.
When there are many documents, reaching zero counts is unlikely, and
thus the sampler is slow to sparsify the topic-concept word matrix
$A$.\footnote{We focus on $A$ in this section because we found that
  $\bar{B}$ is faster to mix; each document may not have many words.
  However, a similar approach could be used to sparsify $B$ as well.}

We introduce an MH procedure to encourage moves of the topic concept-word matrix~$A$ in directions of greater sparsity through joint moves on both~$A$ and~$P$.  Given a proposal distribution~$Q(A',P'\given A,P)$, the acceptance ratio for an MH procedure is given by 
\begin{align*}
  a_{MH} &= 1 \wedge \frac{ p(X\given B,A',P') \,p(A') \, p(P') \,Q(A,P\given A',P')}
   { p(X\given B,A,P) \,p(A)\,p(P) \, Q(A',P'\given A,P)  }
\end{align*}
The sparsity-inducing prior on~$A$ will prefer topic-concept word matrices~$A'$ that have more zeros.  However, as with all Bayesian models (and seen in our case in Equation~\ref{eqn:ll}), when the data get large, the likelihood term~$p( X \given B , A , P )$ will dominate the prior terms~$p( A )$ and~$p( P )$.

To allow for moves toward greater sparsity, our MH proposal uses two core ideas.  First, we use the form of the ontology to propose intelligent split-merge moves for~$A'$.  Second, we attempt to make a move that \emph{keeps the likelihood as constant as possible} by proposing a $P'$ such that ${AP = A'P'}$.  Thus, the prior terms~$p(A)$ and~$p(P)$ will have a larger influence on the move.  The form of~$Q(A',P'\given A,P)$ is as follows:
\begin{itemize}

\item \textbf{$Q(A'\given A,P)$}:  We choose a random topic $k$ and concept word $\cw$.  Let $D_{\cw}$ denote the set of concept words that are descendants of $\cw$ (including $\cw$).  With probability $p_{\sf{split}}$, we sample a random vector~$r$ from $\textrm{Dirichlet}(1_{|D_{\cw}|})$ and create a new $A'_k$ with ${A'_{k\cw} = 0}$ and ${A'_{k\cw'} = A_{k\cw'} + rA_{k\cw}}$, ${\forall \cw'\in D_{\cw}}$. Otherwise, we perform the merge~${A'_{k\cw'} =0}$, ${\forall \cw' \in D_{\cw}}$, and ${A'_{k\cw} = \sum_{\cw' \in D_{\cw} } A_{k\cw'}}$. This split-merge move corresponds to adjusting probabilities in a sub-graph of the ontology, with the merge move corresponding to moving all the mass to a single node.

\item \textbf{$Q(P'\given A',A,P)$}:  Let $P^\star$ be the solution to the optimization problem ${\min_{\hat{P}} ||AP - A'\hat{P}||^2_F}$, where $F$ denotes the Frobenius norm, with the constraints that each row of~$P^\star$ must lie on the simplex and respect the ontology $\mathcal{O}$.  This optimization can be solved as a quadratic program with linear constraints.  We then sample each row of the proposal $P'$ according to~${P'_v \sim \textrm{Dirichlet}( \beta_{MH} P^\star_v )}$. We find in practice that $\beta_{MH}$ generally needs to be large in order to propose appropriately conservative moves.
\end{itemize}
While this procedure can still propose moves over the entire parameter space (thus guaranteeing Harris recurrence on the appropriate stationary distribution corresponding to the prior), it guarantees visits to sparse, high-likelihood solutions with high probability.

\subsection{Adding and Deleting Topics}
Finally, we describe how the number of topics in the data set is automatically learned.  First, we remove any topics that are unused (that is, ${\sum_n\bar{B}_{nk}=0}$).  To propose new topics, we first choose a random document $n$.  We propose a new $A'_{k'}$ from the prior and propose that ${\bar{B'}_{nk'} = 1}$.  Finally, we propose a new ${B'_n \sim \textrm{Dirichlet}( \alpha_B ( \bar{B'}_n\odot {\textstyle \sum_{w}[C_{NKV}]_{n:w}}))}$. The acceptance probability for adding the new topic $A_{k'}$ is given by
\begin{equation*}
  \alpha_{\sf{add}} = 1 \wedge \frac{ p( X_n \given B'_n , A' , P )\, p( B'_n )
    \frac{ \gamma_A }{ N } } { p( X_n \given B_n , A , P ) \, p( B_n ) }
\end{equation*}
where the $\frac{\gamma_A}{N}$ term comes from the probability of adding exactly one new topic in the IBP prior.

\section{Results}

We demonstrate the ability of our Graph-Sparse LDA model to find interpretable, predictive topics on one toy example and two real-world examples from biomedical domains.  In each case we compare our model with the state-of-the-art Bayesian nonparametric topic modeling approach LIDA \cite{archambeau-11}.  We focus on LIDA because it subsumes two other popular sparse topic models, the focused topic model~\cite{williamson-10} and sparse topic model~\cite{wang-09b}, and because the proposed model is a generalization of LIDA.

All samplers were run for 250 iterations.  The topic matrix product~$AP$ was initialized using an LDA tensor decomposition \cite{anandkumar-12} and then factored into $A$ and $P$ using an alternating minimization to find a sparse $A$ that enforced the simplex and ontology constraints.  These initialization procedures reduced the burn-in time.  Finally, a random 1\% of each data-set was held out to compute predictive log-likelihoods.

\paragraph{Demonstration on a Toy Problem}
We first considered a toy problem with a 31-word vocabulary arranged in a binary tree (see Figure~\ref{fig:toy_cartoon}).  There were three underlying topics, each with only a single concept (the three darker nodes in Figure~\ref{fig:toy_cartoon}, labeled 1, 2, and 3).  Each row in the matrix $P_{\cw}$ uniformly distributed 10\% of its probability mass to the ancestors of each concept word and 90\% of its probability mass to the concept word's descendants (including itself).  Each initialization of the problem had a randomly generated document-topic matrix comprising 1000 documents.

\begin{figure*}[tb]
\centering%
\subfloat[][Toy Relative Log-LH]{%
\includegraphics[width=.32\linewidth]{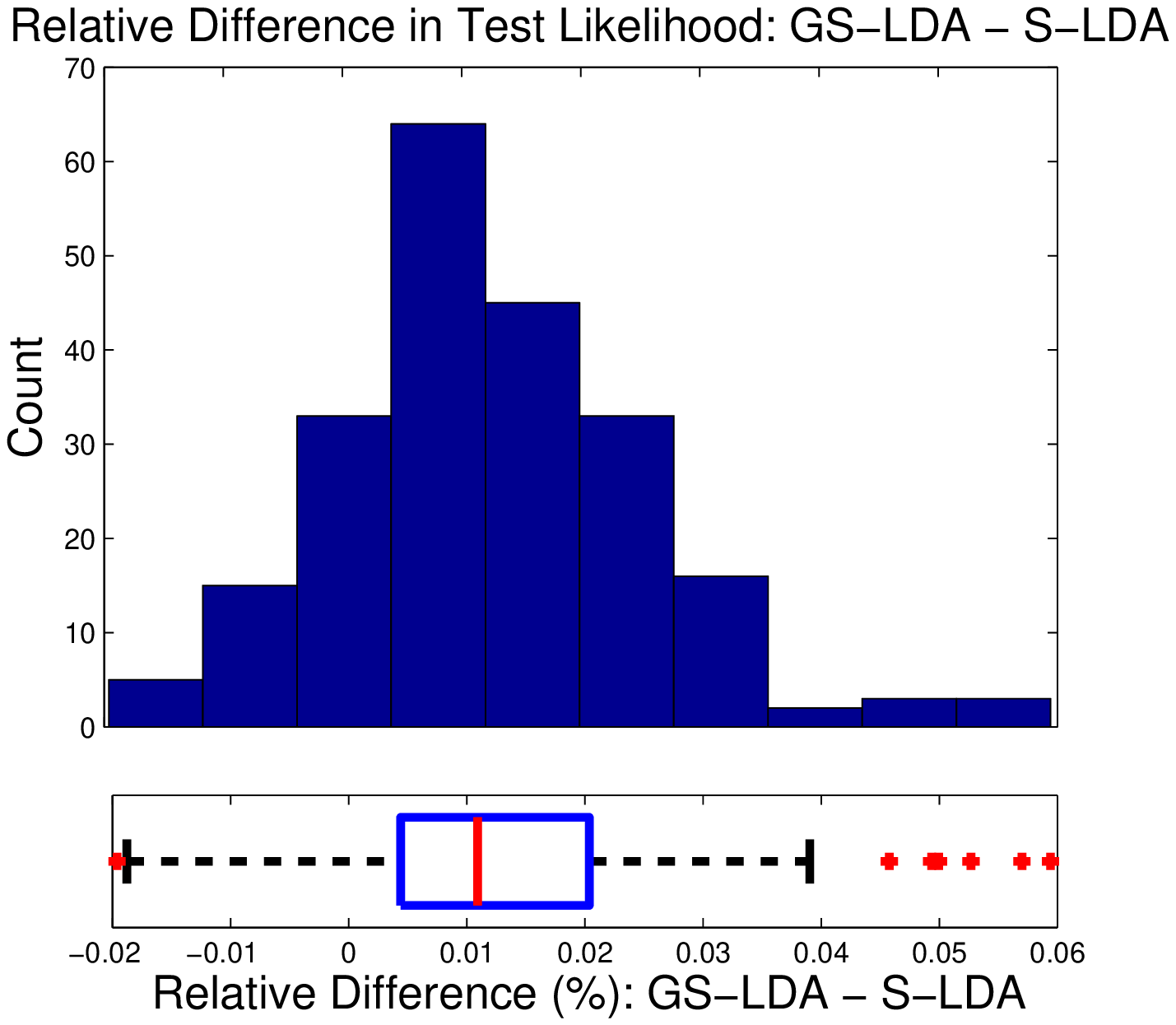}
\label{fig:toy_ll_diff}
}\hfill
\subfloat[Autism Relative Log-LH]{
\includegraphics[width=.32\linewidth]{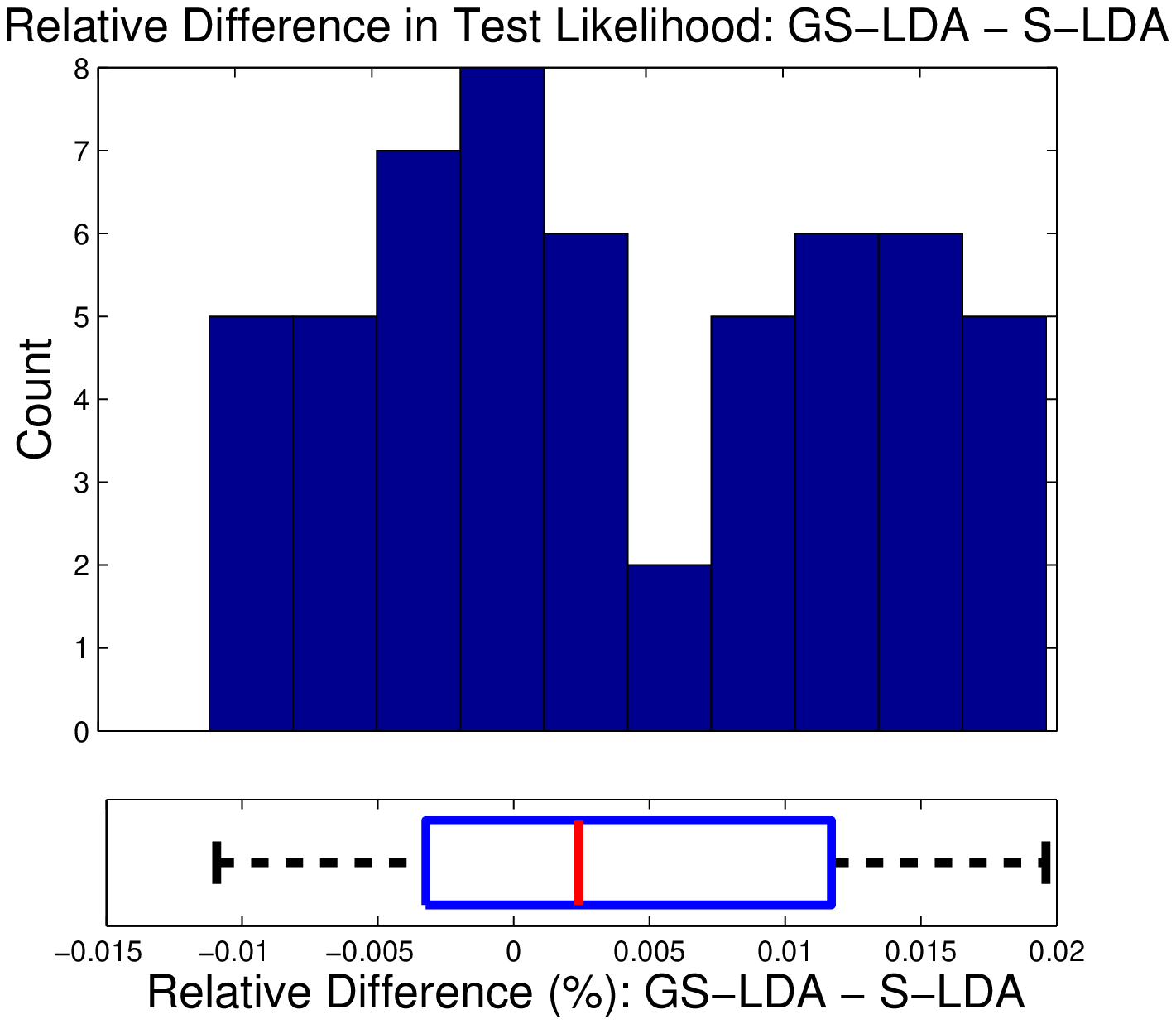}
\label{fig:asd_ll_diff}
}\hfill
\subfloat[SR Relative Log-LH]{
\includegraphics[width=.32\linewidth]{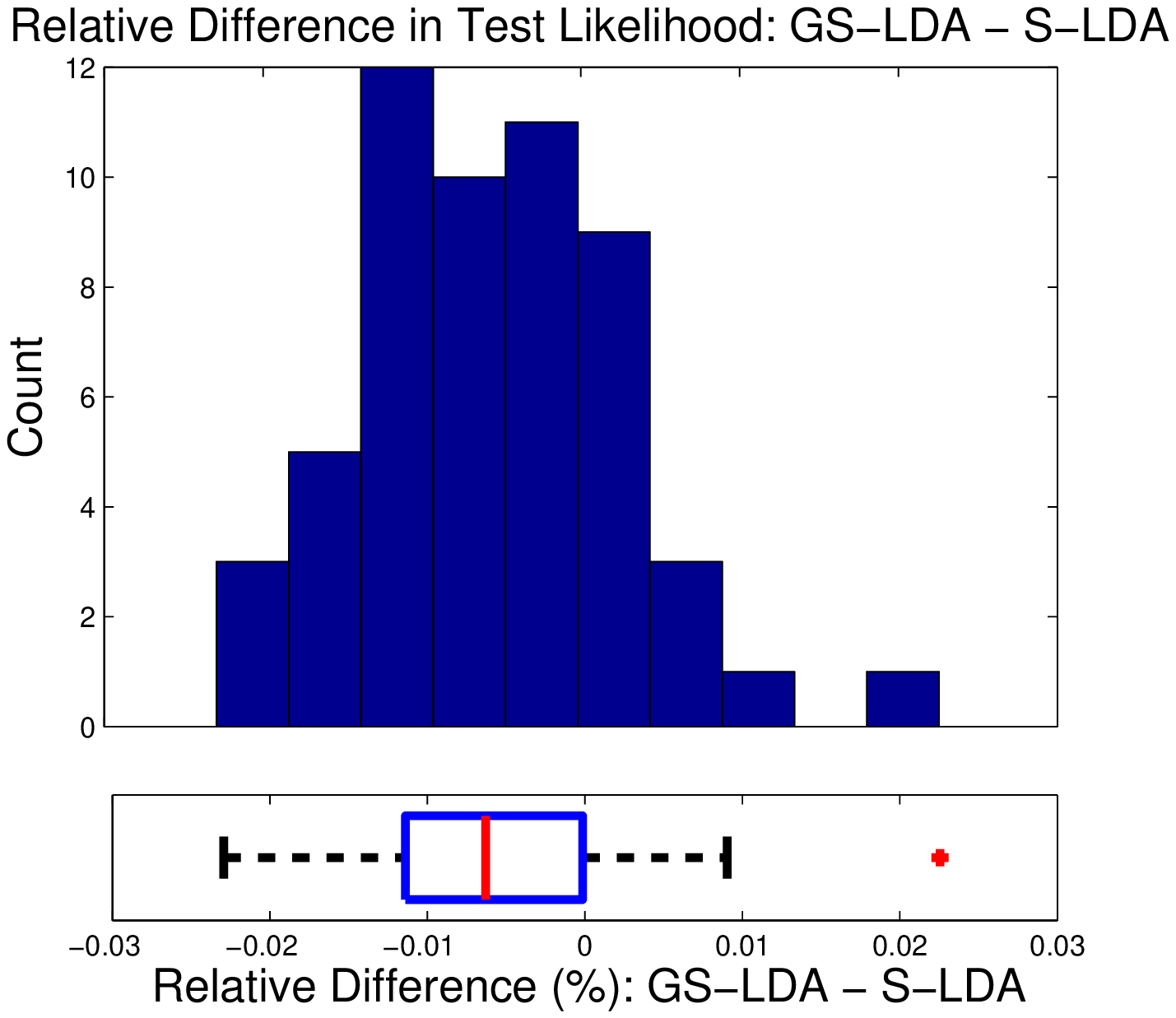}
\label{fig:sr_ll_diff}
}\\%
\subfloat[Toy Topic Sparsity]{
\includegraphics[width=.32\linewidth]{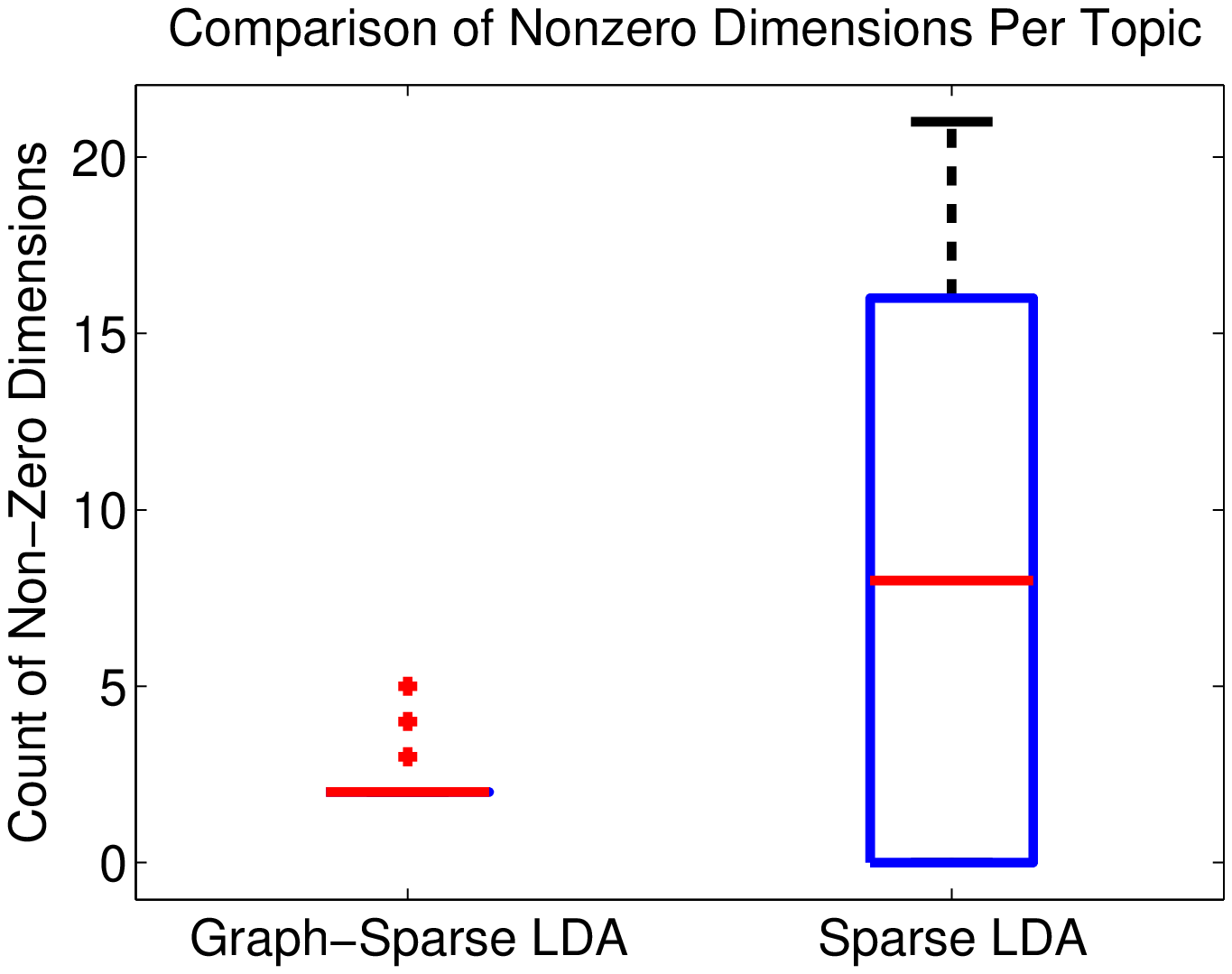}
\label{fig:toy_z_sum}
}\hfill
\subfloat[Autism Topic Sparsity]{
\includegraphics[width=.32\linewidth]{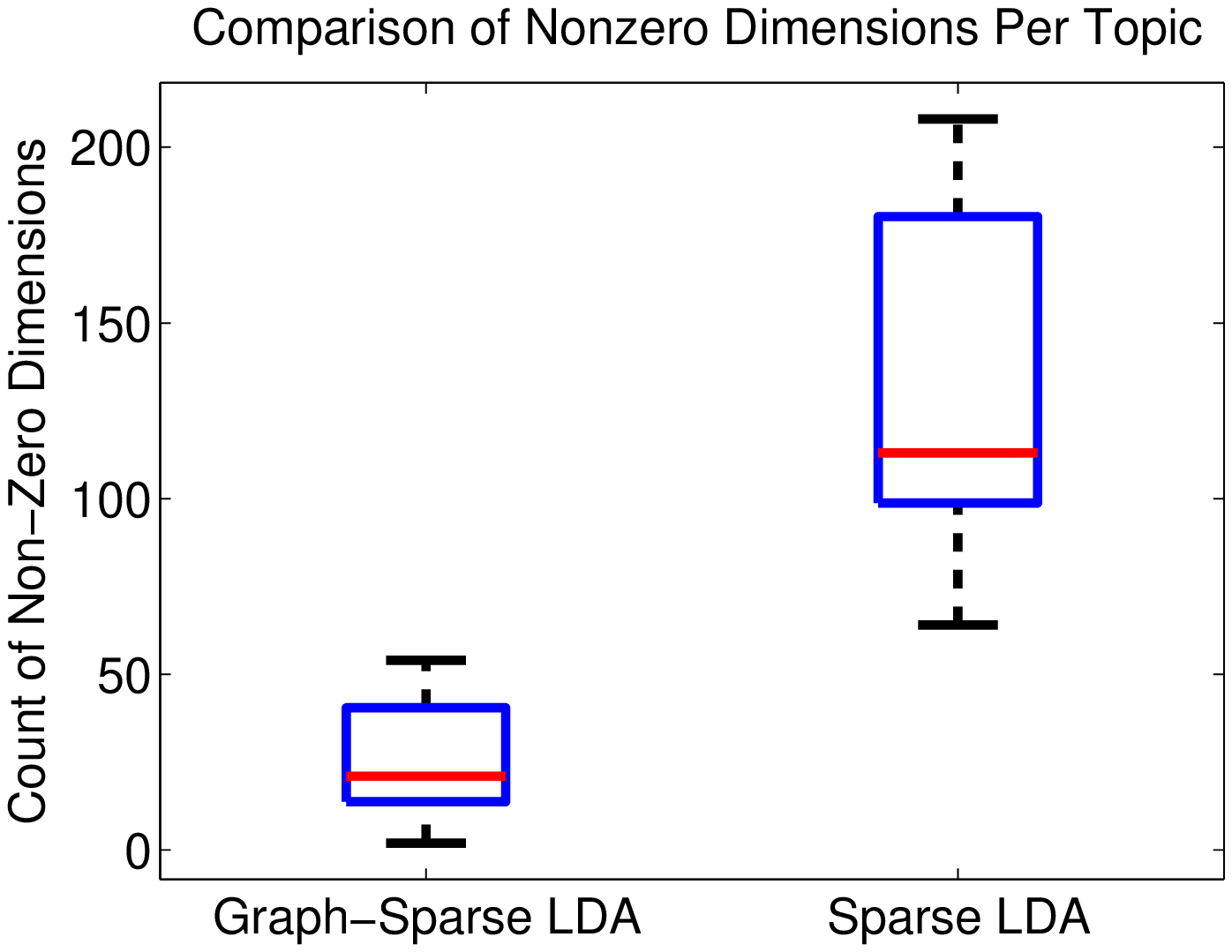}
\label{fig:asd_z_sum}
}\hfill
\subfloat[SR Topic Sparsity]{
\includegraphics[width=.32\linewidth]{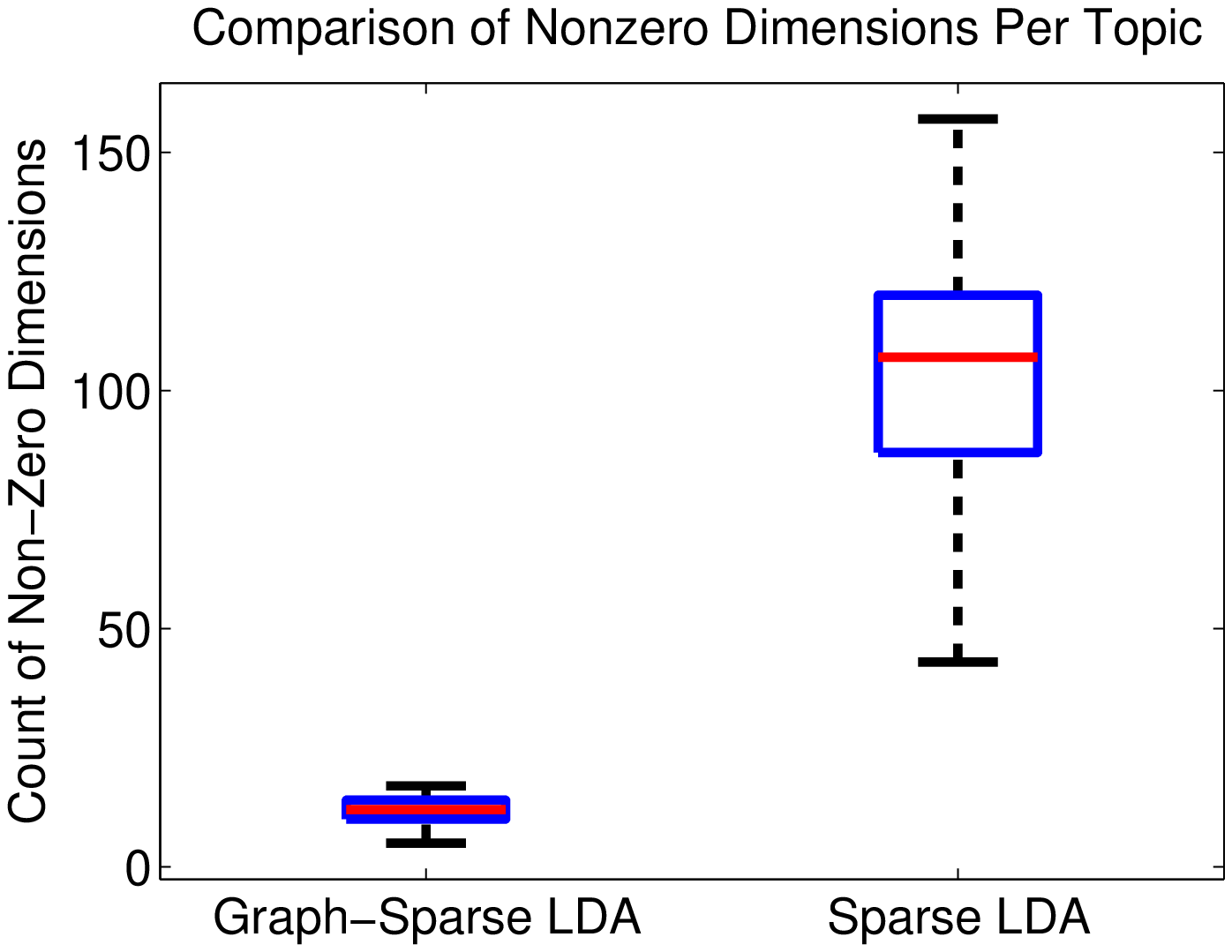}
\label{fig:sr_z_sum}
}
\caption{The top row shows the difference in held-out test
  log-likelihoods between Graph-Sparse LDA and Sparse LDA, divided by
  the overall mean held-out log-likelihood of both models after
  burn-in.  In three domains, the predictive performance of
  Graph-Sparse LDA is within a few percent of LIDA.  The second row
  shows the number of non-zero dimensions in the topic-concept word
  and the topic-word for Graph-Sparse LDA and LIDA models,
  respectively.  Results are shown over 20 independent instantiations
  of the toy problem and 5 independent MCMC runs of the Autism and
  systematic review (SR) problems.}
\label{fig:all_ll}
\vspace{-1.8em}
\end{figure*}

Figures~\ref{fig:toy_ll_diff} and~\ref{fig:toy_z_sum} show the difference in the held-out test likelihoods for the final 50 samples over 20 independent instantiations of the toy problem.  The difference in held-out test likelihoods is skewed positive, implying that Graph-Sparse LDA makes somewhat better predictions than LIDA.  More importantly, Graph-Sparse LDA also recovers a much sparser matrix $A$, as can be seen in figure~\ref{fig:toy_z_sum}. We note, of course, that Graph-Sparse LDA has an additional layer of structure that allows for a very sparse topic concept-word matrix $A$; LIDA does not have access to the ontology information $\mathcal{O}$.  The important point is that by  incorporating this available controlled structured vocabulary into our model, we find a solution with similar or better predictive performance than state-of-the-art models with the additional benefit of a much more interpretable structure.

\paragraph{Patterns of Co-Occurring Diagnoses in Autism Spectrum Disorder}
Autism Spectrum Disorder (ASD) is a complex, heterogenous disease
that is often accompanied by many co-occurring conditions such as
epilepsy and intellectual disability.  We consider a set of 3804
patients with 3626 different diagnoses where the datum $X_{nw}$
corresponds to the number of times patient $n$ received diagnosis
$w$ during the first 15 years of life.\footnote{The Internal Review
  Board of the Harvard Medical School approved this study.} Diagnoses are
organized in a tree-structured hierarchy known as
ICD-9CM \cite{bodenreider-04}.  Diagnoses higher up in the hierarchy
are less specific (such as ``Diseases of the Central Nervous System''
or ``Epilepsy with Recurrent Seizures,'' as opposed to ``Epilepsy,
Unspecified, without mention of intractable epilepsy'').  Clinicians
may encode a diagnosis at any level of the hierarchy, including less
specific ones.

Figure~\ref{fig:asd_ll_diff} shows the difference in test
log-likelihood between Graph-Sparse LDA and LIDA over 5 independent
runs, divided by the overall mean test-likelihood value.  While less
pronounced than in the toy example, Graph-Sparse LDA still has
slightly better predictive performance---certainly on par with current
state-of-the-art topic modeling.  However, the use of the ontology
again allows for much sparser topics, as seen in
Figure~\ref{fig:asd_z_sum}.  In this application, the topics
correspond to possible subtypes in ASD. Being able to concisely
summarize them is the first step toward using the output of this model
for future clinical research.

Finally, Table~\ref{tab:asd_topic} shows an example of one topic
recovered by Graph-Sparse LDA and its corresponding topic discovered
by LIDA.  While the corresponding topic in LIDA has very similar
diagnoses, using the hierarchy allows for Graph-Sparse LDA to
summarize most of the probability mass in this topic in 6 concept
words rather than 119 words.  This topic---which shows a connection
between the more severe form of ASD, intellectual disability, and
epilepsy---as well as the other topics, matched recently published
clinical results on ASD subtypes \cite{doshi-velez-13}.

\begin{table*}[tb]
\centering
\caption{Sample Discovered Topic using Graph-Sparse LDA on the ASD
  data, compared with LIDA.  Graph-Sparse LDA required only 6 concepts
  to summarize most of probability mass in the topic, while LIDA
  required 119.  For LIDA, we do not show all of the diagnoses
  associated with the topic, but only a sample of those diagnoses that
  are summarized by the shown concept words.}
\label{tab:asd_topic}
\resizebox{\textwidth}{!}{%
\begin{tabular}{|p{1.5in}|p{4in}|} \hline
Graph-Sparse LDA & LIDA \\ (6 total nonzero) & (119 total nonzero)
\\ \hline \input{asd_table_clean.txt}
\end{tabular}
}
\end{table*}

\paragraph{Medical Subject Headings for Biomedical Literature}

The National Library of Medicine maintains a controlled structured
vocabulary of Medical Subject Headings (MeSH) \cite{lipscomb-00}.
These terms are hierarchical: terms near the root are more general
than those further down the tree. For example, \emph{cardiovascular
  diseases} subsumes \emph{heart diseases}, which is in turn a parent
of \emph{Heart Aneurysm}.

These MeSH terms are useful for searching the biomedical literature.  For
example, when conducting a \emph{systematic review} (SR)
\cite{grimshaw-93}, one looks to summarize the totality of the
published evidence pertaining to a precise clinical question.
Identifying this evidence in the literature is a time-consuming,
expensive and tedious endeavor; computational methods for reducing the
labor involved in this process have therefore been investigated
\cite{cohen-06,wallace-10}.  MeSH terms are helpful annotations for
facilitating literature screening for systematic reviews, as they can
help researchers undertaking a review quickly decide if
articles are relevant to their query or not.

However, MeSH terms are manually assigned to articles by a small group
of annotators.  Thus, there is inherent variability in the specificity
of the terms assigned to articles.  This variability can make
leveraging the terms difficult. Graph-Sparse LDA provides a means of
identifying latent concepts that define distributions over terms
nearby in the MeSH structure. These interpretable, sparse topics can
provide concise summaries of biomedical documents, thus easing the
evidence retrieval process for overburdened physicians.

\begin{table*}[tb]
\centering
\caption{Sample Discovered Topic using Graph-Sparse LDA on the MeSH
  data for studies comprising the Calcium Channels systematic review,
  compared with LIDA. Superscripts denote the same term found at
  different locations in the MeSH structure; we collapse these when
  they appear sequentially in a topic. Due to space constraints we do
  not show all discovered topics. Graph-Sparse LDA captures the
  concepts ``double-blind trial'' and ``calcium channel blockers'' in
  one topic, which is exactly what the researchers were looking to
  summarize in this systematic review.}
\label{tab:mesh_topic}
\resizebox{\textwidth}{!}{%
\begin{tabular}{|p{1.5in}|p{4.in}|} \hline
Graph-Sparse LDA & LIDA \\ 
(21 total nonzero) & (90 total nonzero) \\ \hline
\emph{0.565}: Double-Blind Method$^{1,2,3,4}$ & (\textbf{1}) \emph{0.353}: Double-Blind Method$^{1,2,3,4}$ \\ \hline 
\emph{0.110}: Calcium Channel Blockers$^{1,2}$ & (\textbf{7}) \emph{0.031} Adrenergic beta-Antagonists$^{1}$, \emph{0.026} Drug Therapy, Combination, \emph{0.022} Calcium Channel Blockers, \emph{0.016} Felodipine, \emph{0.015} Atenololm, \emph{0.006} Benzazepines, \emph{0.01} Mibefradil$^{1,2}$  \\ \hline
\emph{0.095}: Angina Pectoris$^{1,2}$ & (\textbf{3}) \emph{0.030}: Angina Pectoris$^{2}$, \emph{0.030}: Myocardial Ischemia$^{1,2}$, \emph{0.003}: Atrial Flutter \\ \hline
\end{tabular}
}
\end{table*}

We consider a dataset of 1218 documents annotated with 5347 unique
MeSH terms (23 average terms per document) that were screened for a
systematic review of the effects of calcium-channel blocker (CCB)
drugs \cite{cohen-06}.  In figure~\ref{fig:sr_ll_diff}, we see that
the test log-likelihood for Graph-Sparse LDA on these data is on par
with LIDA, while producing a much sparser summary of concept-words
(figure~\ref{fig:sr_z_sum}).  Here, the concepts found by Graph-Sparse
LDA correspond to sets of MeSH terms that might help researchers
rapidly identify studies reporting results for trials investigating
the use of CCB's---without having to make sense of a topic comprising
hundreds of unique MeSH terms.

Table~\ref{tab:mesh_topic} shows the top concept-words in a sample
topic discovered by Graph-Sparse LDA compared to a similar topic
discovered by LIDA.  Graph-Sparse LDA gives most of topic mass to
double-blind trials and CCBs; knowing the relative prevalence in an
article of this topic would clearly help a researcher looking to find
reports of randomized controlled trials of CCBs.  In contrast, words
related to concept CCBs are divided among terms in LIDA.  Some of the
LIDA terms, such as \emph{Drug Therapy, Combination} and
\emph{Mibefradil} are also present in Graph-Sparse LDA, but with much
lower probability -- the concept CCB summarizes most of the instances.
We note that a professional systematic reviewer at [Anonymous] confirmed
that the more concise topics found by Graph-Sparse LDA would be more
useful in facilitating evidence retrieval tasks than those found by
LIDA.

\section{Discussion and Related Work}
\label{section:related-work}

Topic models \cite{blei-03,steyvers-07} have gained wide popularity as
a flexible framework for uncovering latent structure in corpora.
Existing topic models have typically assumed that observed words are
unstructured. By contrast, here we have considered scenarios in which these
words are drawn from a known underlying structure (such as an ontology).

Prior work in interpretable topic models has focused on various
notions of coherence.  \cite{tealeaves} introduced the idea of
``intrusion detection'' where they hypothesized that a more coherent,
or interpretable, topic would be one where a human annotator would be
able to identify an inserted ``intruder'' word among the top 5 words
in a topic; \cite{ml_tealeaves} automated this process.  Contrary to
expecation, they found that interpretability (as rated by human
annotators) was \emph{negatively} correlated with test likelihood.
\cite{coherence_newman} and \cite{coherence_mimno} developed measures
of topic coherence that strongly correlated with human annotations of
topic quality.  

However, the evaluations in all of these works still
focus only on the top $n$ words in a topic (which powerfully indicates
how linked sparsity is to interpretability; humans have
trouble working with long lists).  In contrast, our approach does not
sacrifice on predictive quality and, by using the ontological
structure, provides a compact summary that describes \emph{most} of
the words, not just the top $n$.  This quality is particularly
valuable in the kinds of scenarios that we described, where annotation
disagreement or diagnostic ``slosh'' can result in a large number of
words with non-trivial probabilities.

This use of a human-provided structure to induce interpretability also
distinguishes Graph-Sparse LDA from other hierarchical and tree
structured topic models where the structure is typically learned.  For
example, \cite{blei-10} use a nested Chinese Restaurant Process to
learn hierarchies of topics where subtopics are more specific than
their parents.  \cite{chen-11} expand on this idea with a
nonparametric Markov model that allows a subtopic to have multiple
parents.  \cite{hu-12} develop inference techniques for sparse
versions of these tree-structured topic models.  Learned hierarchies
have also been used to capture correlations between topics, such as
\cite{li-06}.  In all of these models, the learned hiearchical
structure allows for various kinds of statistical sharing between the
topics.  However, each topic is still a distribution over a large
vocabulary, and the interpretation task is only complicated by
requiring a human to now both inspect the hierarchy and the topics for
structure.

Among the fully unsupervised approaches, the closest to our work is
the super-word concept modeling of \cite{el-12}, which uses a nested
Beta process to describe a document with a sparse set of super-words,
or concepts, each of which are associated with a sparse set of words.
Known auxiliary information about the words, encoded in a feature
vector, can be used to encourage or discourage words from being part
of the same concept.  A key difference in our approach is that we use
the graph structure to guide the formation of concepts, which
maintains interpretability while removing the need for each concept to
have a sparse set of words.  Our graph-structured relationships also
result in a much simpler inference procedure.

While not applied to increase interpretability, expert-defined
hierarchies have been used in topic models in other contexts.  Early
work by \cite{abney-99} used hierarchies for word-sense
disambiguation in n-gram tuples. This idea was later incorporated into a
topic modeling context by \cite{boyd-graber-07}.  Other work has used
hierarchical structure as partial supervision to improve
topic-modeling output in scenarios in which some words come from
controlled vocabularies (or have known relationships) and others do
not.  \cite{slutsky-13} consider representing the content of website
summaries via a hierarchical model. Their approach exploits
ontological structure by jointly modeling word and ontology term
generation. They showed that their model (which leverages the
hierarchical document labels) improved on existing approaches with
respect to perplexity.  \cite{andrzejewski-09} use Dirichlet forest
priors to enforce expert-provided ``must be in same topic'' and
``cannot be in same topic'' constraints between words.  Finally,
\cite{perotte-11} propose a hierarchically supervised LDA model where
there is a hierarchy on the document labels (rather than on the
vocabulary).  Specifically, they treat categories as `labels' and
model the assignment of these to documents via (probit) regression
models. Their model stipulates that when a node is assigned to a
category so too are all of its parents, thus capturing the
hierarchical structure.  In contrast to all of these works, which
focus on prediction tasks, Graph-Sparse LDA uses the ontology in a
probabilistic --- rather than enforced --- manner to obtain sparse topics
from extant controlled vocabularies.

We note that our word generation model is much more
general than other approaches.  Here we have considered scenarios in which
the ontological structure allows a concept-word to generate words that
are its descendants and ancestors.  However, we can imagine that a
concept-word can generate any nearby observed word, where the
definition of ``nearby'' is entirely up to the model-designer
(concretely, ``nearby'' corresponds to the sparsity pattern in the
matrix $P$).  This difference allows for much more flexibility in
modeling: the underlying structure can be a tree, a DAG, or just some
collection of neighborhoods.  At the same time, our formulation
results in a Gibbs sampling procedure that is simpler than many other
hierarchical models.

\section{Conclusions}
Topic models have revolutionized prediction and classification models in
many domains, and many scientists are now attempting to use them to
uncover structure from their data.  For these applications, however,
prediction is not enough: scientists wish to be able to
\emph{understand} the structure in order to posit new theories. At the same
time, structured knowledge-bases often exist for scientific domains; these
are information dense resources that capture a wealth of expertise. 

In this paper we have proposed a model that exploits such resources to
achieve the stated aim of identifying interpretable topics.  More
specifically, we have described a novel Bayesian nonparametric model,
Graph-Sparse LDA, that leverages existing controlled vocabulary
structures to induce interpretable topics. The Bayesian nonparametric
aspect of the model allows us to discover the number of topics in our
dataset. Leveraging ontological knowledge allows us to uncover sparse
sets of concept words that provide succinct, interpretable topic
summaries that maintain the ability to explain a large number of
observed words.  The combination of this representational power and an
efficient inference procedure allowed us to realize topic
interpretability while still matching (and often exceeding)
state-of-the-art predictive performance.

While we have focused on controlled vocabularies in the biomedical domain,
this approach could be more generally applied to text corpora using
standard hierarchies such as WordNet \cite{miller-95}.  In these more
general domains, using hierarchies could eliminate the need for basic
pre-processing such as stemming.  This model is relatively straight-forward to
implement, and we expect it to be useful for a variety of topic or
factor-discovery applications where the observed dimensions have some
human-understandable relationships.

\section*{Acknowledgments} We are grateful to Isaac Kohane and the i2b2 team at Boston Children's Hospital for providing us the autism data and their feedback on the GS-LDA model as a data-mining tool.
 
\bibliography{arxiv} \bibliographystyle{ieeetr}

\end{document}

%% file: asd_table_clean.txt
\emph{0.333}: Autistic disorder, current or active state & \textbf{(1)} \emph{0.213}: Autistic disorder, current or active state \\ \hline 
\emph{0.203}: Epilepsy and recurrent seizures & \textbf{(15)}, including \emph{0.052}: Epilepsy, unspecified, without mention of intractable epilepsy, \emph{0.0283}: Localization-related epilepsy and epileptic syndromes with com, \emph{0.023}: Generalized convulsive epilepsy, without mention of intractable epilepsy, \emph{0.008}: Localization-related epilepsy and epileptic syndromes with sim, \emph{0.006}: Generalized convulsive epilepsy, with intractable epilepsy, \emph{0.005}: Epilepsy, unspecified, with intractable epilepsy, \emph{0.004}: Infantile spasms, without mention of intractable epilepsy, ... \\ \hline
\emph{0.131}: Other convulsions & \textbf{(2)} \emph{0.083}: Other convulsions, \emph{0.015}: Convulsions \\ \hline 
\emph{0.055}: Downs syndrome & \textbf{(1)} \emph{0.001}: Conditions due to anomaly of unspecified chromosome\\ \hline 
\emph{0.046}: Intellectual disability & \textbf{(1)} \emph{0.034}: Intellectual disability \\ \hline 
\emph{0.040}: Other Disorders of the Central Nervous System & \textbf{(31)}, including: \emph{0.052}: Epilepsy, unspecified, without mention of intractable epilepsy, \emph{0.006}: Generalized convulsive epilepsy, with intractable epilepsy, \emph{0.002}: Other brain condition, \emph{0.002}: Quadriplegia, \emph{0.0001}: Hemiplegia, unspecified, affecting dominant side, \emph{0.0001}: Migraine without aura, with intractable migraine, \emph{0.00009}: Flaccid hemiplegia Flaccid hemiplegia and hemiparesis affecting unspecified side, \emph{0.00005}: Metabolic encephalopathy... \\ \hline